%% file: article.tex
\documentclass[12pt,english,a4paper]{article}
\usepackage[T1]{fontenc}
\usepackage[utf8]{inputenc}
\usepackage{babel}
\usepackage{algorithm}
\usepackage{algorithmic}
\usepackage{graphicx}
\usepackage{epsfig}
\usepackage{amsmath}
\usepackage{fullpage}
\usepackage{cmlgc}
\usepackage{amsthm}
\usepackage{setspace}
\usepackage{amsfonts}
\usepackage{rotating}
\usepackage{setspace}
\usepackage{color}

\graphicspath{./fig/}

\def\ie{{\it i.e.}~}

\renewcommand{\v}[1]{\mathbf{#1}}

\newcommand{\takeout}[1]{}

\graphicspath{fig/}

\newcommand{\citeref}[2]{\vspace{-#1} {\em #2} \vspace{#1} }

\author{I\~{n}aki Fern\'{a}ndez Pérez\\Université de Lorraine, LORIA
\\ {\tt inaki.fernandez@loria.fr}
\\Amine Boumaza\\Université de Lorraine, LORIA\\
{\tt amine.boumaza@loria.fr}
\\Fran\c{c}ois Charpillet\\INRIA Nancy Grand-Est, LORIA
\\ {\tt francois.charpillet@loria.fr}
\\Campus scientifique BP 239 Vandoeuvre-lès-Nancy Cedex, F-54506,}

\title{Comparison of Selection Methods in On-line Distributed
  Evolutionary Robotics}

\begin{document}

\maketitle \citeref{140mm}{Pérez, I. F., Boumaza, A. and Charpillet, F. Comparison of Selection Methods in On-line Distributed Evolutionary Robotics. In Proceedings of the Int. Conf. on the Synthesis and Simulation of Living Systems (Alife'14), pages 282-289, MIT Press, 2014.}

\abstract{In this paper, we study the impact of selection methods in the
  context of on-line on-board distributed evolutionary algorithms.  We
  propose a variant of the mEDEA algorithm in which we add a selection
  operator, and we apply it in a task-driven scenario. We evaluate
  four selection methods that induce different intensity of selection
  pressure in a multi-robot navigation with obstacle avoidance task
  and a collective foraging task. Experiments show that a small
  intensity of selection pressure is sufficient to rapidly obtain good
  performances on the tasks at hand. We introduce different measures
  to compare the selection methods, and show that the higher the
  selection pressure, the better the performances obtained, especially
  for the more challenging food foraging task.}

\input{01-intro}
\input{02-related}
\input{03-algorithms}
\input{04-experiments}

\input{05-conclusion}

\bibliographystyle{plain}
\bibliography{refs}

\end{document}

%% file: 01-intro.tex
\section{Introduction}
\label{sec:intro}
Evolutionary robotics (ER) (\cite{Nolfi-2000}) aims to design robotic
agents' behaviors using evolutionary algorithms (EA)
(\cite{Eiben-2003}). In this context, EA's are traditionally seen as
a tool to optimize agents' controllers w.r.t to an explicit objective
function (fitness). This process is carried out in an off-line
fashion; once the behavior is learned and the controller optimized,
the agents are deployed and their controllers' parameters remain
fixed.

On the other hand, on-line evolution (\cite{Watson2002}) takes a
different approach in which behavior learning is performed during the
actual execution of a task. In these algorithms, learning or
optimization is a continuous process, \ie robotic agents are
constantly exploring new behaviors and adapting their controllers to
new conditions in their environment. Usually, this is referred to as
adaptation.

These two visions of ER can be related to on-line and off-line
approaches in Machine Learning (ML). Off-line ML algorithms learn a
specific task and solutions should generalize to unseen situations
after the learning process, whereas on-line ML algorithms
progressively adapt solutions to new presented situations while
solving the task. In this sense, both on-line ML algorithms and
on-line EA's perform lifelong adaptation or learning, to possibly
changing environments or objectives.

In this paper, we focus on on-line distributed evolution of swarm
robotic agents. We are interested in learning individual agents
behaviors in a distributed context where the agents adapt their
controllers to environmental conditions independently while
deployed. These agents may locally communicate with each other and do
not have a global view of the swarm. In this sense, this approach
finds many ties with Artificial Life, where the objective is to design
autonomous organisms that adapt to their environment.

On-line distributed evolution may be viewed as distributing an EA on
the swarm of agents. Traditional evolutionary operators (mutation,
crossover etc.) are performed on the agents and local communication
ensures the spread of genetic material in the population of agents.

In EA's, selection operators drive evolution toward fit individuals
by controlling the intensity of selection pressure to solve the given
task. These operators and their impact on evolutionary dynamics have
been extensively studied in off-line contexts (\cite{Eiben-2003}). In
this paper, we study their impact in on-line distributed ER, where
evolutionary dynamics are different to the off-line case:
selection is performed locally on partial populations and fitness
values on which selection is performed are not reliable. Our
experiments show that, in this context, a strong selection pressure
leads to the best performances, contrary to classical approaches in
which lower selection pressure is preferred, to maintain diversity in
the population. This result suggests that, in distributed ER
algorithms, diversity is already maintained by the disjoint
sub-populations.

%The property of local selection in these algorithms implies that the
%individuals that survive through evolution in different agents are
%different, thus inherently maintaining a certain degree of diversity
%over the whole population of controllers in the swarm. As such, as
%opposed to classical centralized ER approaches, elitist selection
%operators in distributed ER, which induce a high intensity of
%selection pressure, can lead to good individuals, even in the absence
%of explicit mechanisms to ensure diversity, as our experiments show.

%Furthermore, as local populations are built based on communication
%which, in turn, depends on the positions of the agents, their
%behaviors influence the diversity of the population over the swarm.
%Furthermore, the sub-population of an agent is built upon local
%communication and thus depend on the agents' behaviors. This means
%that a part of the exploration (in the optimization sense) is
%delegated to the agents' behaviors which can reduce the need of
%diversity maintaining mechanisms at the EA level.

Several authors have addressed on-line evolution of robotic agent
controllers in different contexts: adaptation to dynamically changing
environments (\cite{dinu2013}), parameter tuning (\cite{eiben2010}),
evolution of self-assembly (\cite{bianco2004}), communication
(\cite{Pineda-2012}), phototaxis and navigation
(\cite{Karafotias2011}, \cite{Silva2012}). Some of this work is
detailed in the next section. The authors use different selection
mechanisms inducing different intensities of selection pressure to
drive evolution. In this paper, we compare different selection
operators and measure the impact they have on the performances of
learning two swarm robotics tasks: navigation with obstacle avoidance
and collective food foraging.

We begin by reviewing different selection schemes proposed in the
context of on-line distributed ER and then we present the algorithm
that will serve as a test bed, along with the selection methods we
compare. In the fourth section, we detail our experimental setting
and discuss the results. Finally, we close with some concluding
remarks and future directions of research.

% Local IspellDict:        english

%% file: 02-related.tex
\section{Related Work}\label{sec:rw}

In the following, we review several on-line distributed ER algorithms
and discuss the selection mechanisms that were applied to ensure the
desired intensity of selection pressure in order to drive evolution.

A common characteristic of on-line distributed ER algorithms is that
each agent has one controller at a time, that it executes (the active
controller), and locally spreads altered copies of this controller to
other agents. In this sense, agents have only a partial view of the
population in the swarm (a local repository). Fitness assignment or
evaluation of individual chromosomes is performed by the agents
themselves and is thus noisy, as different agents evaluate their
active controllers in different conditions. Selection takes place when
the active controller is to be replaced by a new one from the
repository.

% PGTA 
PGTA (Probabilistic Gene Transfer Algorithm) introduced
by \cite{Watson2002}, is commonly cited as the first implementation of
a distributed on-line ER algorithm. This algorithm evolves the weights
of fixed-topology neural controllers and agents exchange parts (genes)
of their respective chromosomes using local broadcasts. The algorithm
considers a virtual energy level that reflects the performance of the
agent's controller. This energy level increases every time the agents
reach an energy source and decreases whenever communication takes
place. Furthermore, the rate at which the agents broadcast their genes
is proportional to their energy level and conversely, the rate at
which they accept a received gene is inversely proportional to their
energy level. This way, selection pressure is introduced in that fit
agents transmit their genes to unfit ones.

% odNEAT
\cite{Silva2012} introduced odNEAT, an on-line distributed version of
NEAT (Neuro-Evolution of Augmenting Topologies) (\cite{Stanley2002}),
where each agent has one active chromosome that is transmitted to
nearby agents. Collected chromosomes from other agents are stored in a
local repository within niches of species according to their
topological similarities, as in NEAT. Each agent has a virtual energy
level that increases when the task is performed correctly and
decreases otherwise. This energy level is sampled periodically to
measure fitness values and, whenever this level reaches zero, the
active chromosome is replaced by one in the repository. At this point,
a species is selected based on its average fitness value, then a
chromosome is selected within this species using binary
tournament. Each agent broadcasts its active chromosome at a rate
proportional to the average fitness of the species it belongs
to. This, added to the fact that the active chromosome is selected
from fit niches, maintains a certain selection pressure toward fit
individuals.
 
% EDEA
EDEA (Embodied Distributed Evolutionary Algorithm)
(\cite{Karafotias2011}), was applied to different swarm robotics
tasks: phototaxis, navigation with obstacle avoidance and collective
patrolling. In this algorithm, each agent possesses one chromosome,
whose controller is executed and evaluated on a given task. At each
iteration, agents broadcast their chromosomes alongside with their
fitness to other nearby agents with a given probability (fixed
parameter). Upon reception, an agent selects a chromosome from those
collected using binary tournament. This last chromosome is then
mutated and recombined (using crossover) with the current active
chromosome with probability $\frac{f(x')}{s_c \times f(x)}$, where
$f(x')$ is the fitness of the selected chromosome, $f(x)$ is the
fitness of the agent's current chromosome and $s_c$ is a scalar
controlling the intensity of selection pressure. To ensure an accurate
measure of fitness values, agents evaluate their controllers for at
least a minimum period of time (maturation age), during which agents
neither transmit nor receive other chromosomes.

%MEDEA 
With mEDEA (minimal Environment-driven Distributed Evolutionary
Algorithm), \cite{medea2010} address evolutionary adaptation with
implicit fitness, \ie without a task-driven fitness function. The
algorithm takes a gene perspective in which successful chromosomes are
those that spread over the population of agents and which requires:
1) to maximize mating opportunities and 2) to minimize the risk for
agents (their vehicles).

At every time step, agents execute their respective active controllers
and locally broadcast mutated copies of the corresponding
chromosomes. Received chromosomes (transmitted by other agents) are
stored in a local list. At the end of the execution period
(\textit{lifetime}), the active chromosome is replaced with a randomly
selected one from the agent's list and the list is emptied. An agent
dies if there are no chromosomes in its list (if it did not meet other
agents) and it remains dead until it receives a chromosome from
another agent passing by.

The authors show that the number of living agents rises with time and
remains at a sustained level. Furthermore, agents develop navigation
and obstacle avoidance capabilities that allow them to better spread
their chromosomes. This work shows that environment-driven selection
pressure alone can maintain a certain level of adaptation in a swarm
of robotic agents. A slightly modified version of this algorithm is
used in this work and is detailed in the next section.

%MONEE
\cite{MONEE2013a} proposed MONEE (Multi-Objective aNd open-Ended
Evolution), an extension to mEDEA adding a task-driven pressure as
well as a mechanism (called market) for balancing the distribution of
tasks among the population of agents, if several tasks are to be
tackled. Their experiments show that MONEE is capable of improving
mEDEA's performances in a collective concurrent foraging task, in
which agents have to collect items of several kinds.

The authors show that the swarm is able to adapt to the environment
(as mEDEA ensures), while foraging different kinds of items
(optimizing the task-solving behavior). In this context, each type of
item is considered a different task. The algorithm uses an explicit
fitness function in order to guide the search toward better performing
solutions. The market mechanism, which takes into account the scarcity
of items, ensures that agents do not focus on the most frequent kind
of items (the easiest task), thus neglecting less frequent ones. In
their paper, the agent's controller is selected using rank-based
selection from the agent's list of chromosomes. The authors argue that
when a specific task is to be addressed, a task-driven selection
pressure is necessary. This idea is discussed in the remainder of this
paper.

% END
In the aforementioned works, authors used different classical
selection operators from evolutionary computation in on-line
distributed ER algorithms. It is however not clear if these operators
perform in the same fashion as when they are used in an off-line
non-distributed manner. In an on-line and distributed context,
evolutionary dynamics are different, since selection is performed
locally at the agent level and over the individuals whose vehicles had
the opportunity to meet. In addition, and this is not inherent to
on-line distributed evolution but to many ER contexts, fitness
evaluation is intrinsically noisy as the agents evaluate their
controllers in different conditions, which may have a great impact on
their performance. A legitimate question one could ask is: does it
still make sense to use selection?

In this paper, we compare different selection methods corresponding to
different intensities of selection pressure in a task-driven context.
We apply these methods in a modified version of mEDEA and measure
their impact on two different swarm robotics tasks.

% Local IspellDict:        english

%% file: 03-algorithms.tex
\section{Algorithms}
\label{sec:algo}
In this section, we describe the variant of mEDEA we used in our
experiments (Algorithm~\ref{alg:VarmEDEA}). It is run by all the
agents of the swarm independently in a distributed manner. At any
time, each agent possesses a single controller which is randomly
initialized at the beginning of evolution.

The main difference w.r.t. mEDEA is that the algorithm alternates
between two phases, namely an evaluation phase, in which the agent
runs, evaluates and transmits its controller to nearby listening
agents, and a listening phase, in which the agent does not move and
listens to incoming chromosomes, sent by nearby agents. The evaluation
and the listening phases last $T_{\mathrm{e}}$ and $T_{\mathrm{l}}$
respectively, and, for different robots, they take place at different
moments. Since the different robots are desynchronized, robots in the
evaluation phase are able to spread their genomes to other robots that
are in the listening phase.

If only one common phase takes place, an agent that turns on the spot
transmits its controller to any fitter agent crossing it, as broadcast
and reception are simultaneous. This separation in two phases is
inspired from MONEE where it is argued that it lessens the spread of
poorly achieving controllers. Also, task-driven selection was
introduced in MONEE to simultaneously tackle several tasks.

The agent's controller is executed and evaluated during the evaluation
phase. For each agent, this phase lasts $T_{\mathrm{e}}$ time-steps at
most\footnote{A little random number is substracted from
  $T_{\mathrm{e}}$ so as the evaluation phases of the agents are not
  synchronized.}. During this phase, at each time-step the agent
executes its current controller by reading the sensors' inputs and
computing the motors' outputs. The agent also updates the fitness
value of the controller, based on the outcome of the its actions, and
locally broadcasts both the chromosome corresponding to its controller
and its current fitness value.

Once the $T_{\mathrm{e}}$ evaluation steps are elapsed, the agent
begins its listening phase, which lasts $T_{\mathrm{l}}$
time-steps. During this phase, the agent stops and listens for
incoming chromosomes from nearby passing agents (agents that are in
their evaluation phase). These chromosomes are transmitted along with
their respective fitness values. Consequently, at the end of this
phase, an agent has a local list of chromosomes and fitnesses, or
local population. Another difference w.r.t mEDEA is that the local
population also contains the agent's current chromosome. This is done
to ensure that all agents always have at least one chromosome in their
respective populations, which happens particularly when an agent is
isolated during its listening phase and does not receive any other
chromosome. In mEDEA, isolated agents stay inactive until they receive
a chromosome from another agent passing by.

\begin{algorithm}[h]
 \caption{\label{alg:VarmEDEA}mEDEA}
 \begin{algorithmic}[1]
    \STATE {$g_a := random()$}
    \WHILE{true}  
    \STATE {$\v{l} := \emptyset $}
    \STATE {// Evaluation phase}
    \FOR{$t = 1$ to $T_{\mathrm{e}}$} 
	\STATE{$exec(g_a)$}
	\STATE{$broadcast(g_a)$}
    \ENDFOR
    \STATE {// Listening phase}
    \FOR{$t = 1$ to $T_{\mathrm{l}}$} 
        \STATE $\v{l} := \v{l} \bigcup listen()$
    \ENDFOR
    \STATE{$\v{l} := \v{l} \bigcup \{g_a\}$}
    \STATE $selected := select(\v{l})$
    \STATE $g_a := mutate(selected)$
    \ENDWHILE
  \end{algorithmic}
\end{algorithm}

After the listening period, the agent needs to load a new controller
for its next evaluation phase. To do so, it selects a chromosome from
its list using one of the selection methods discussed further. The
selected chromosome is then mutated and becomes the agent's active
controller. In this case, mutation consists in adding a normal random
variable with mean $0$ and variance $\sigma^2$ to each gene (each
synaptic weight of the neuro-controller).

Once the next controller is chosen, the list is emptied.  This means
selection is performed on a list of chromosomes that have been
collected by the agent during the previous listening phase. At this
time, the new controller's evaluation phase begins. We consider one
iteration of the algorithm (evaluation plus listening phase) as one
generation.

The selection method selects the new chromosome among the collected
ones based on their fitness. This can be done in different manners,
depending on the desired intensity of selection pressure. In this
paper we compare four different selection methods, each one defining a
different intensity of task-driven selection pressure. The choice of
these selection methods aims at giving a large span of intensities of
selection pressure, from the strongest (\textit{Best}), to the lowest
(\textit{Random}):

 \paragraph{Best Selection:}
 This method deterministically selects the controller with the highest
 fitness. This is the selection method with the strongest selection
 pressure, as the agent will never be allowed to select a controller
 with a lower fitness than the previous one. \textit{Best} selection
 can be compared to an elitist selection scheme where the best fit
 controllers are always kept.
 \begin{algorithm}[h] 
    \caption{\label{alg:SelBest}Best selection}
    \begin{algorithmic}[1]
      \STATE order $x_i$ and index as $x_{i:n}$ such that:\\$f(x_{1:n}) \ge f(x_{2:n}) \ge \ldots \ge f(x_{n:n})$  
      \STATE {\bf return} $x_{1:n}$  
    \end{algorithmic}
  \end{algorithm}
  
\paragraph{Rank-Based Selection:}
In this case, selection probabilities are assigned to each controller
according to their rank, \ie the position of the controller in the
list, once sorted w.r.t. fitness values. The best controller has the
highest probability of being selected; however, less fit controllers
still have a positive probability of being selected. Traditionally,
this method is preferred to Roulette Wheel selection that assigns
individuals probabilities proportional to their fitness values, which
highly biases evolution toward best individuals.

\begin{algorithm}[h]
  \caption{\label{alg:SelRB}Rank-based selection}
  \begin{algorithmic}[1]
    \STATE order $x_i$ and index as $x_{i:n}$ such that:\\$f(x_{1:n}) \ge f(x_{2:n}) \ge \ldots \ge f(x_{n:n})$ 
    \STATE select $x_{i:n}$ with probability $Pr(x_{i:n})=\frac{n+1-i}{1+2+\ldots+n}$ 
    \STATE {\bf return} $x_{i:n}$  
  \end{algorithmic}
\end{algorithm}
 
\paragraph{Binary Tournament:}
This method uniformly samples a number of controllers equal to the
size of the tournament (two in our case) and selects the one with the
highest fitness. Here, the selection pressure is adjusted through the
size of the tournament: the higher the size, the higher the selection
pressure, the extreme case being when the tournament size is equal to
the size of the population. In this case, the best controller is
chosen\footnote{It is assumed that sampling is performed without
  replacement.}. Conversely, when the size of the tournament is two,
the induced selection pressure is the lowest.

\begin{algorithm}[h]
  \caption{\label{alg:SelTour}$k$-Tournament selection}
  \begin{algorithmic}[1]
    \STATE uniformly sample $k$ $x_i$, noted $\{x_{1:k},\ldots,x_{k:k}\}$
    \STATE order $x_{i:k}$ such that:\\ $f(x_{1:k}) \ge f(x_{2:k}) \ge \ldots \ge f(x_{k:k})$
    \STATE {\bf return} $x_{1:k}$
  \end{algorithmic}
\end{algorithm}

\paragraph{Random Selection:}
This method selects a controller in the local population at random,
disregarding its fitness value and therefore inducing no task-driven
selection pressure at all. \textit{Randon} selection is considered as
a baseline for comparisons with the other methods that effectively
induce a certain task-driven selection pressure. As discussed in the
previous section, this is the selection operator used by mEDEA for
evolving survival capabilities of the swarm without any task-driven
explicit goal. By considering \textit{Random} in our experiments, we
aim to compare the original mEDEA selection scheme with more selective
operators.

Each one of these four selection methods induces a different intensity
of selection pressure on the evolution of the swarm. In the next
section, we describe our experiments comparing the impact of each one
of these intensities.

% Local IspellDict:        english

%% file: 04-experiments.tex
\section{Experiments}
\label{sec:exp}
We compare these selection methods on a set of experiments in
simulation for two different tasks, fast-forward navigation and
collective foraging, which are two well-studied benchmarks in swarm
robotics. All our experiments were performed on the RoboRobo
simulator~(\cite{roborobo}).
 
\subsection{Description}
\begin{figure}[!h]
  \centering
  \resizebox{.55\linewidth}{!}{\includegraphics{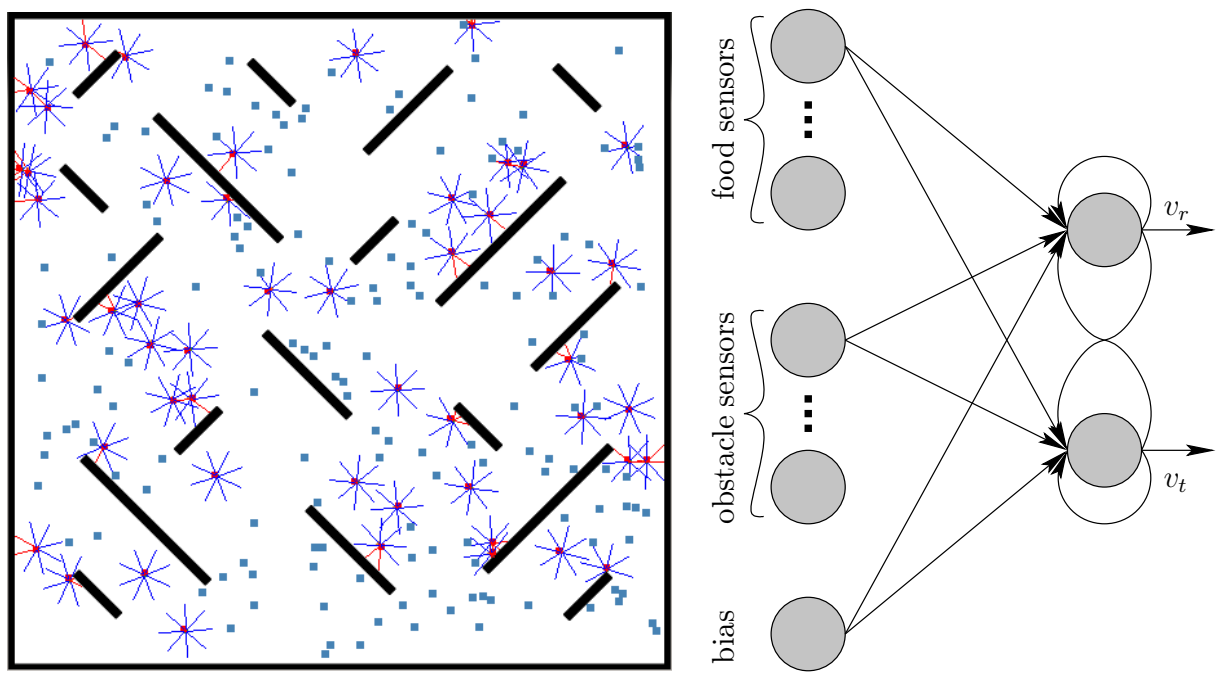}}\hfill
  \resizebox{.42\linewidth}{!}{\input{fig/controller.pstex_t}}
  \caption{\label{fig:envir} Left: the simulation environment
    containing agents (red dots with thin hairlines representing
    sensors), obstacles (dark lines) and food items (blue
    dots). Right: the architecture of the neuro-controller.}
\end{figure}

In all experiments, a swarm of robotic agents is deployed in a bounded
environment containing static obstacles (black lines in
Figure~\ref{fig:envir}). Agents also perceive other agents as
obstacles.

All the agents in the swarm are morphologically homogeneous, \ie they
have the same physical properties, sensors and motors, and only differ
in the parameters of their respective controllers. Each agent has $8$
obstacle proximity sensors evenly distributed around the agent, and
$8$ food item sensors are added in the case of the foraging task. An
item sensor measures the distance to the closest item in the direction
of the sensor. These simulated agents are similar to Khepera or e-puck
robots.

We use a recurrent neural network as the architecture of the
neuro-controllers of the agents (Figure~\ref{fig:envir}). The inputs
of the network are the activation values of all sensors and the 2
outputs correspond to the translational and rotational velocities of
the agent. The activation function of the output neurons is a
hyperbolic tangent, taking values in $[-1,+1]$. Two bias connections
(one for each output neuron), as well as $4$ recurrent connections
(previous speed and previous rotation for both outputs) are
added. This setup yields $22$ connection weights for the navigation
task and $38$ for the foraging task in the neuro-controller. The
chromosome of the controller is the vector of these
weights. Table~\ref{tab:param} summarizes the different parameters
used in our experiments.

In the navigation task, agents must learn to move as fast and straight
as possible in the environment while avoiding obstacles, whereas in
the foraging task, agents must collect food items present in the
environment (Figure~\ref{fig:envir}). An item is collected when an
agent passes over it, at which time it is replaced by another item at
a random location.

We define the fitness function for the navigation task after the one
introduced in (\cite{Nolfi-2000}). Each agent $r$ computes its fitness
at generation $g$ as:
\begin{equation}
f_r^g = \sum_{t = 1}^{T_e} v_{t}(t) \cdot (1 - |v_r(t)|) \cdot
min(a_s(t))
\end{equation}
where $v_{t}(t)$, $v_r(t)$ and $a_s(t)$ are respectively the
translational velocity, the rotational velocity and the activations of
the obstacle sensors of the agent at each time-step $t$ of its
evaluation phase.

In the foraging task, a controller's fitness is computed as the number
of items collected during its evaluation phase. Furthermore, since we
are interested in the performance of the entire swarm, we define the
swarm fitness as the sum of the individual fitness of all agents at
each generation:
\begin{equation}
F_s(g) = \sum_{r \in swarm} f_r^g
\end{equation}

\begin{table}
  \centering
  \caption{\label{tab:param} Experimental settings.}
  \begin{tabular}{c c}
    \hline
    \multicolumn{2}{c}{Experiments}\\
    \hline
    Number of food items &  150\\
    Swarm size & 50 agents\\
    Exp. length & $5 \times 10^5$ sim. steps\\
    Number of runs & 30\\
    \hline 
    \multicolumn{2}{c}{Evolution} \\
    \hline 
    Evolution length & $\sim$ 250 generations \\
    $T_e$ & $2000 - rand(0,500)$ sim. steps \\
    $T_l$ & 200 sim. steps \\
    Chromosome size & Nav.: 22, Forag.: 38 \\
    Mutation step-size & $\sigma = 0.5$ \\\hline 
  \end{tabular}
\end{table}

% Measures 
\subsection{Measures}
A characteristic of on-line ER is that agents learn as they are
performing the actual task in an open-ended way. In this context, the
best fitness ever reached by the swarm is not a reliable measure,
since it only reflects a ''good'' performance at one point of the
evolution.  Furthermore, fitness evaluation is inherently noisy, due
to different evaluation conditions encountered by the
agents. Therefore, we introduce four measures that will be used to
compare the impact of the different selection methods. These measures
summarize information on the swarm spanning over several generations.
They are used only for evaluation and comparison of the selection
methods and are computed once the evolution has ended. A pictorial
description of these four measures is shown in
Figure~\ref{fig:metrics}.
\begin{figure}[!h]
  \centering
  \resizebox{.49\linewidth}{!}{\input{fig/measure-1.pstex_t}}
  \resizebox{.49\linewidth}{!}{\input{fig/measure-2.pstex_t}}
  \resizebox{.49\linewidth}{!}{\input{fig/measure-3.pstex_t}}
  \resizebox{.49\linewidth}{!}{\input{fig/measure-4.pstex_t}}
  \caption{\label{fig:metrics}A pictorial description of the four
    comparison measures. From top to bottom and left to right: the
    average accumulated swarm fitness, the fixed budget swarm fitness,
    the time to reach target and the accumulated fitness above target.}
\end{figure}
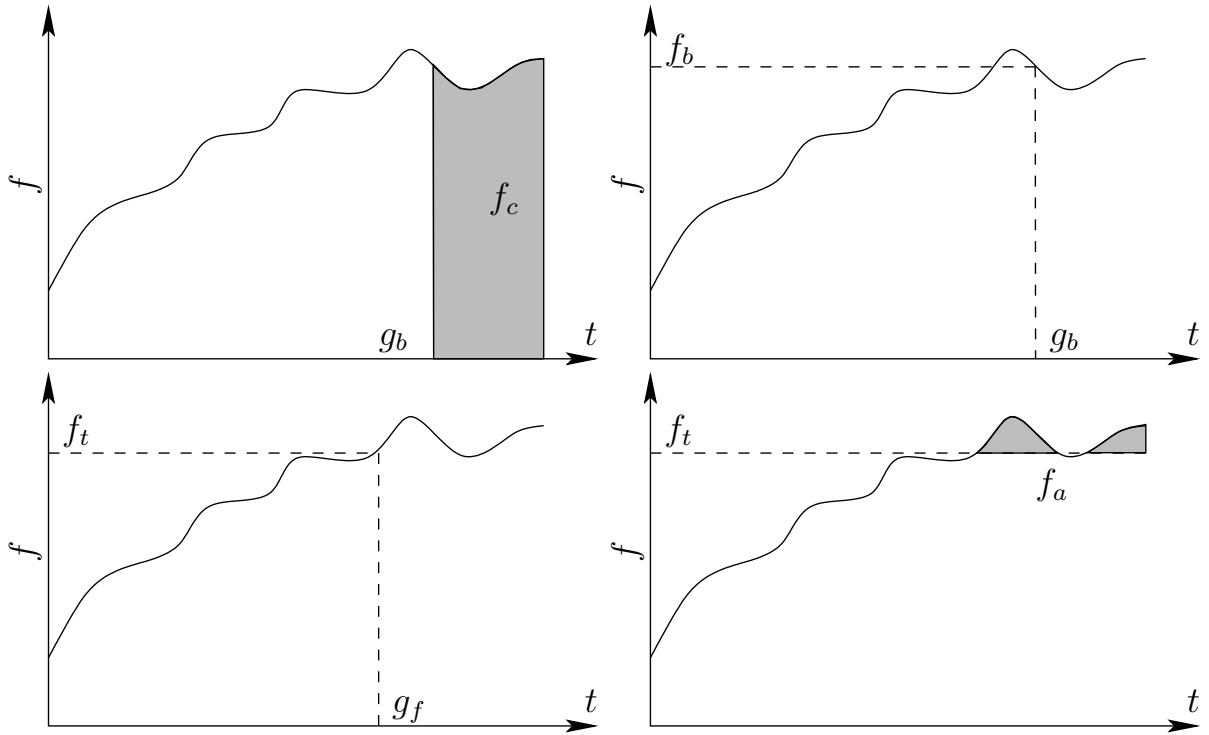
\begin{itemize}
\item Average accumulated swarm fitness ($f_c$): is the average swarm
  fitness in the last generations. This metric reflects the
  performance of the swarm at the end of the evolution. In our
  experiments, we compute the average over the last $8\%$ generations.
\item Fixed budget swarm fitness ($f_b$): is the swarm fitness
  reached at a certain generation (computational budget). This measure
  helps to compare different methods on the same grounds. In our
  experiments, we measure this value at $92\%$ of the evolution, which
  corresponds to the first generation considered in the computation of
  $f_c$.
\item Time to reach target ($g_f$): is the first generation at which a
  predefined target fitness is reached. If this level is never
  reached, $g_f$ corresponds to the last generation. We fixed the
  target at $80\%$ of the maximum fitness reached over all runs and
  all selection methods. This metric reflects a certain convergence
  rate of the algorithms, \ie how fast the swarm hits the target
  fitness on the task at hand.
\item Accumulated fitness above target ($f_a$): is the sum of all
  swarm fitness values above a predefined target fitness. It reflects
  to which extent the target level is exceeded and if this performance
  is maintained over the long run. We used the same target fitness as
  with $g_f$.
\end{itemize}
These comparison measures are not to be taken individually.  For
instance $f_c$ and $f_b$ complement each other and give an indication
of the level and stability of the performance reached by the swarm at
the end of evolution. If $f_b$ and $f_c$ are close then performance of
the swarm is stable. Also, $g_f$ and $f_a$ combined reflect how
fast a given fitness level is reached and to which extent that level
is exceeded. Adding the two latter measures to $f_c$ shows if that
trend is maintained.

%Since $F_s$ includes the fitness values of all agents in the swarm, it
%corresponds to global information which is not available to each agent
%during evolution. Consequently, we use these metrics to evaluate and
%compare the results of evolution using different selection methods,
%and they are not used by the algorithms themselves to guide learning.

\subsection{Results and discussion}
\begin{figure}[!h]
  \centering
  \resizebox{.95\linewidth}{!}{\includegraphics[width=\textwidth,angle=-90,origin=c]{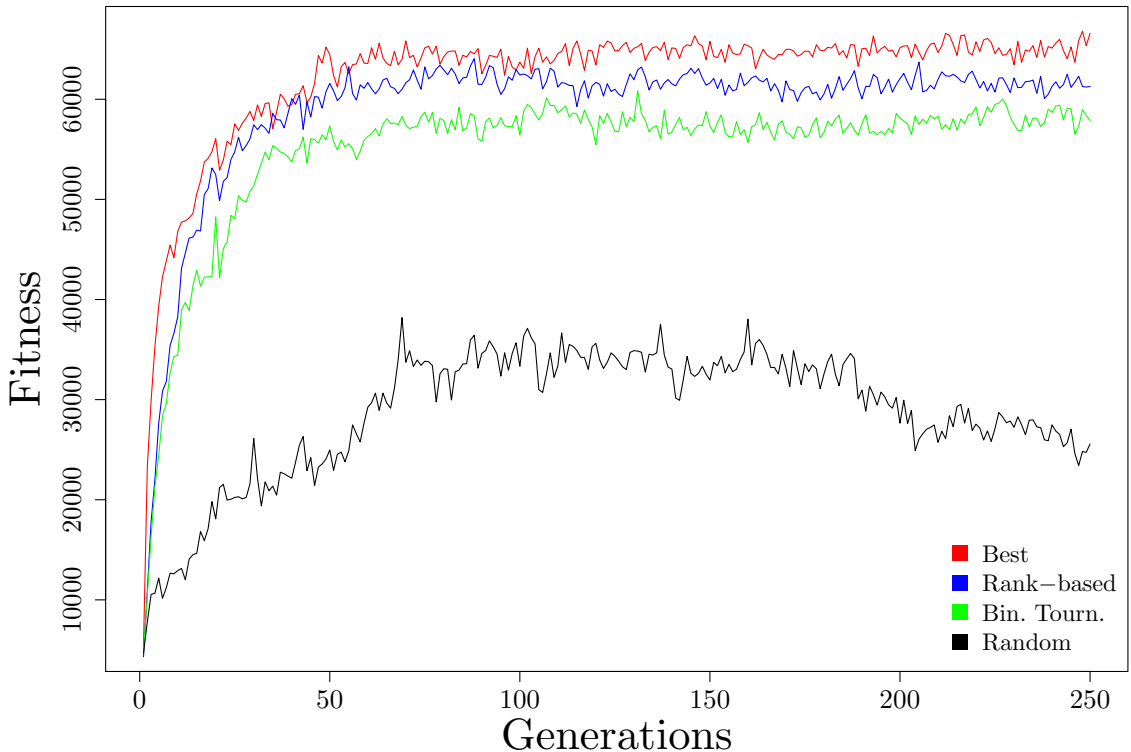}}
  \vspace{-35px}
   \caption{\label{fig:navigEvo} Median swarm fitness per generation
     over the $30$ runs for the navigation task.}
\end{figure}

\begin{figure}[!h]
  \centering
   \resizebox{.95\linewidth}{!}{\includegraphics[width=\textwidth,angle=-90,origin=c]{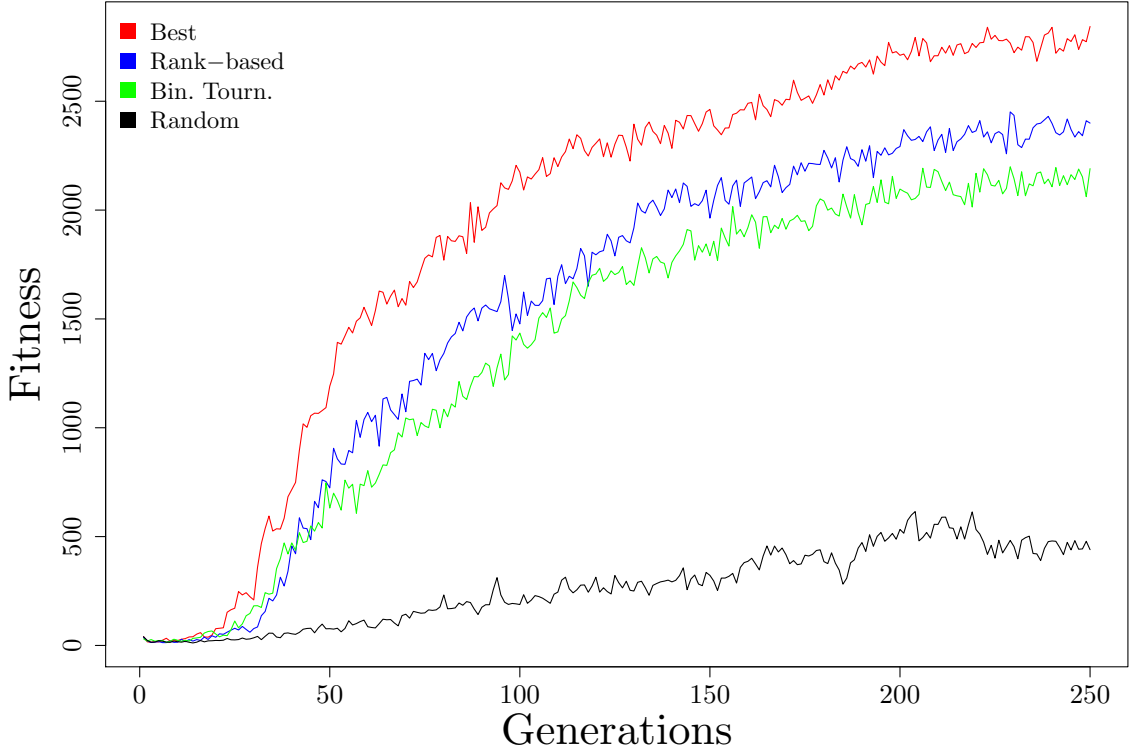}}
   \vspace{-35px}
   \caption{\label{fig:foragEvo} Median swarm fitness per generation
     over the $30$ runs for the foraging task.}
\end{figure}

\begin{figure}[!h]
  \centering
  \resizebox{.49\linewidth}{!}{
    \includegraphics[width=\textwidth,angle=-90,origin=c]{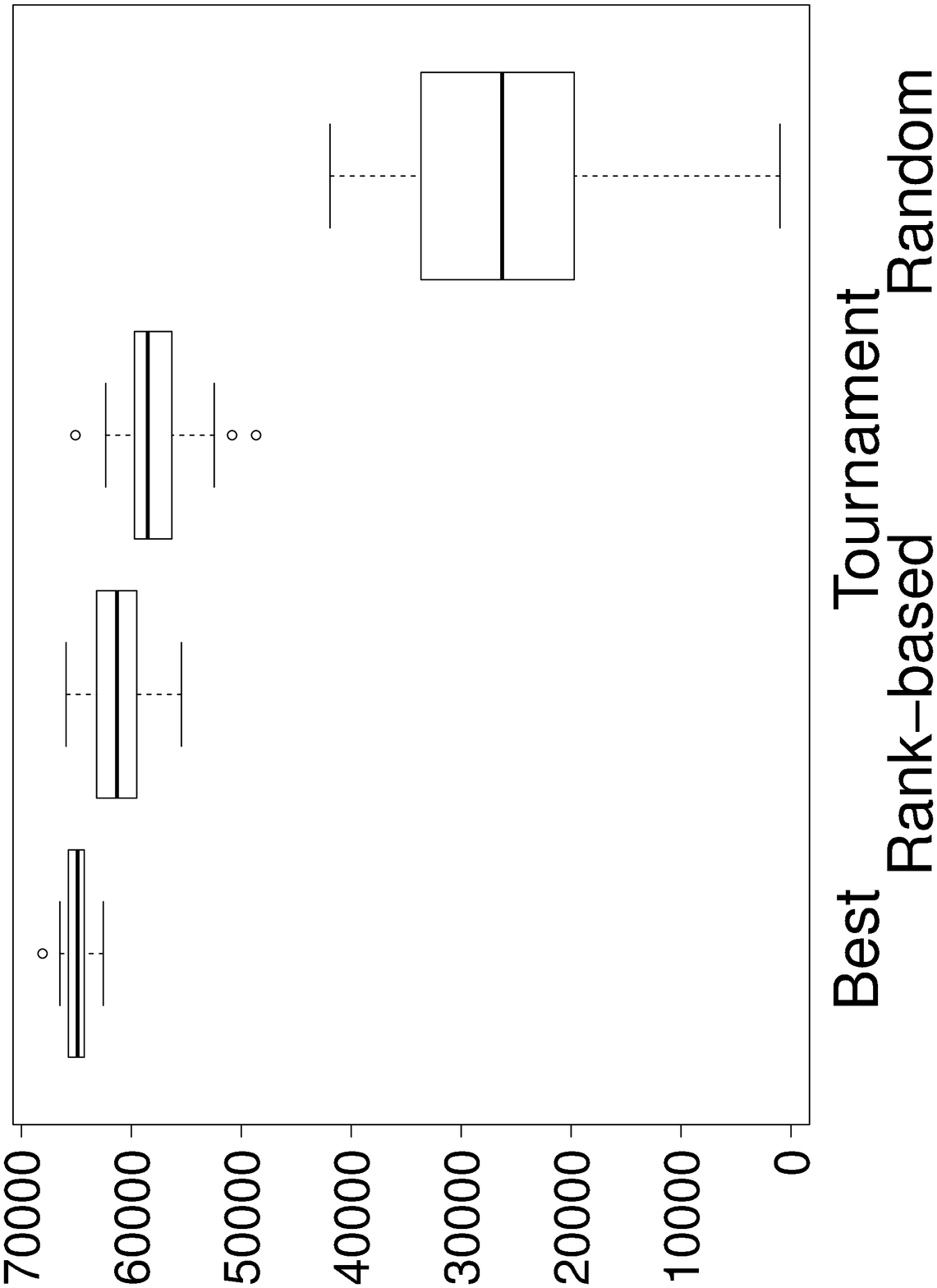}}
  \resizebox{.49\linewidth}{!}{
    \includegraphics[width=\textwidth,angle=-90,origin=c]{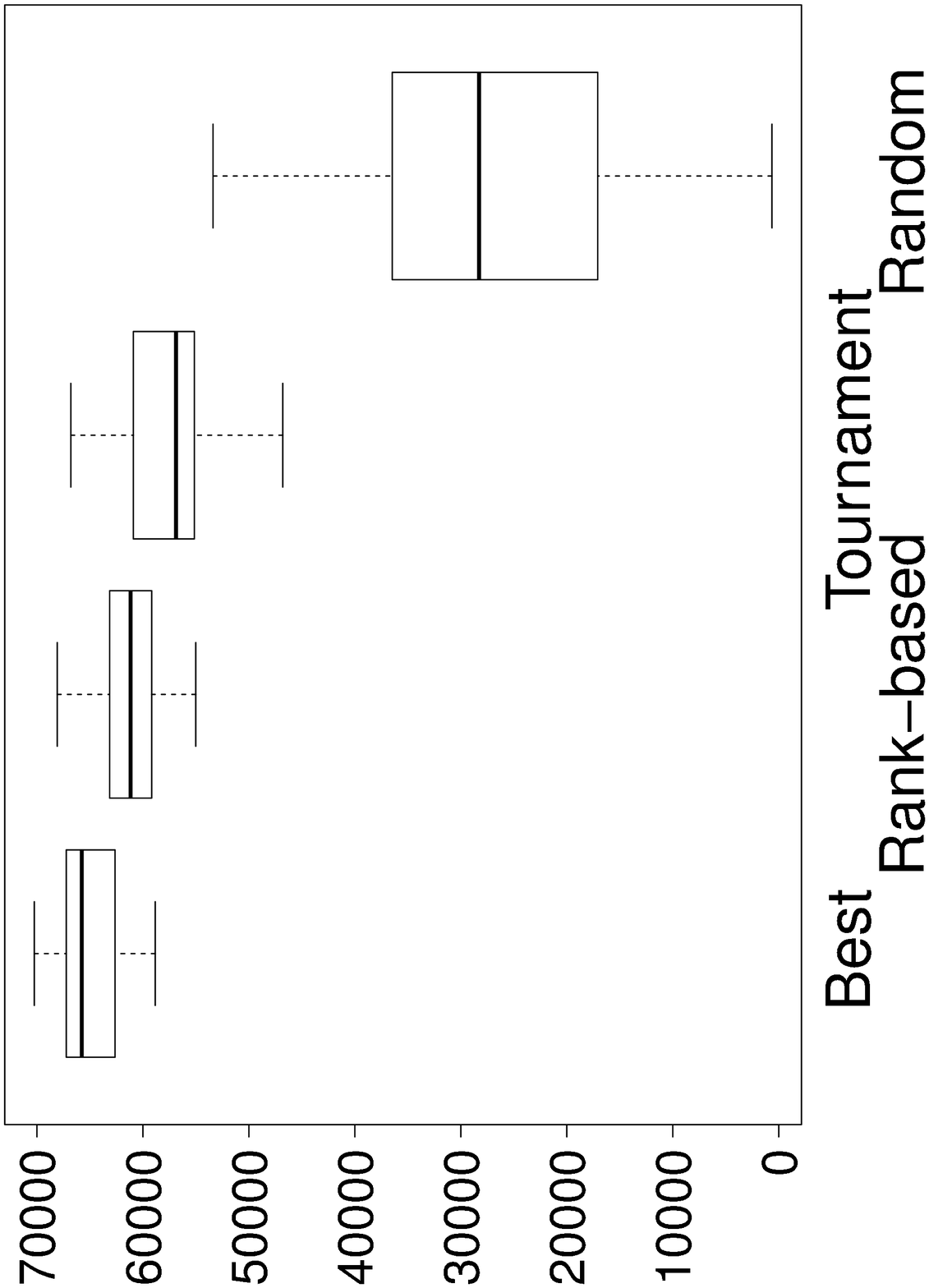}}
  \resizebox{.49\linewidth}{!}{
    \includegraphics[width=\textwidth,angle=-90,origin=c]{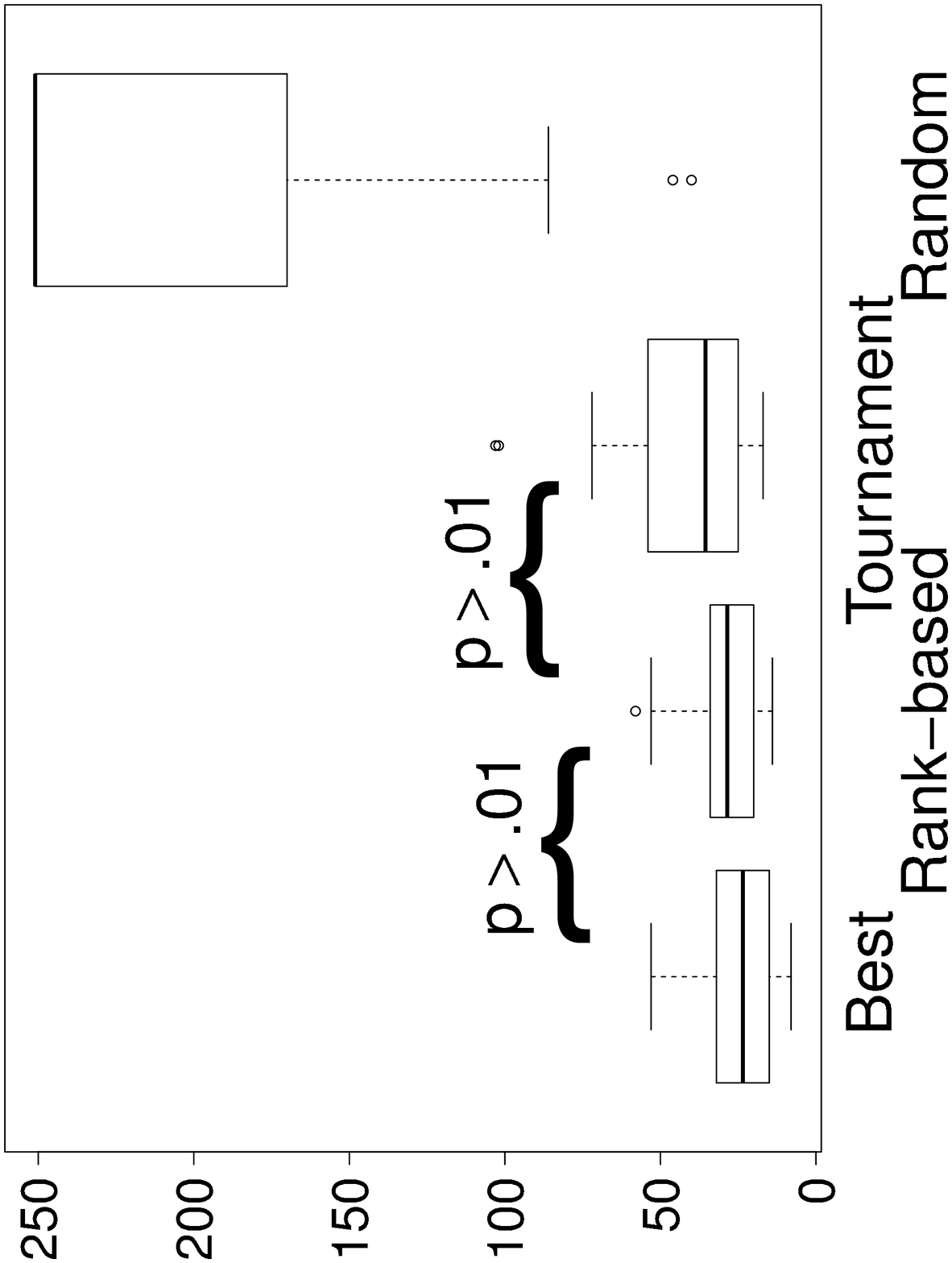}}
  \resizebox{.49\linewidth}{!}{
    \includegraphics[width=\textwidth,angle=-90,origin=c]{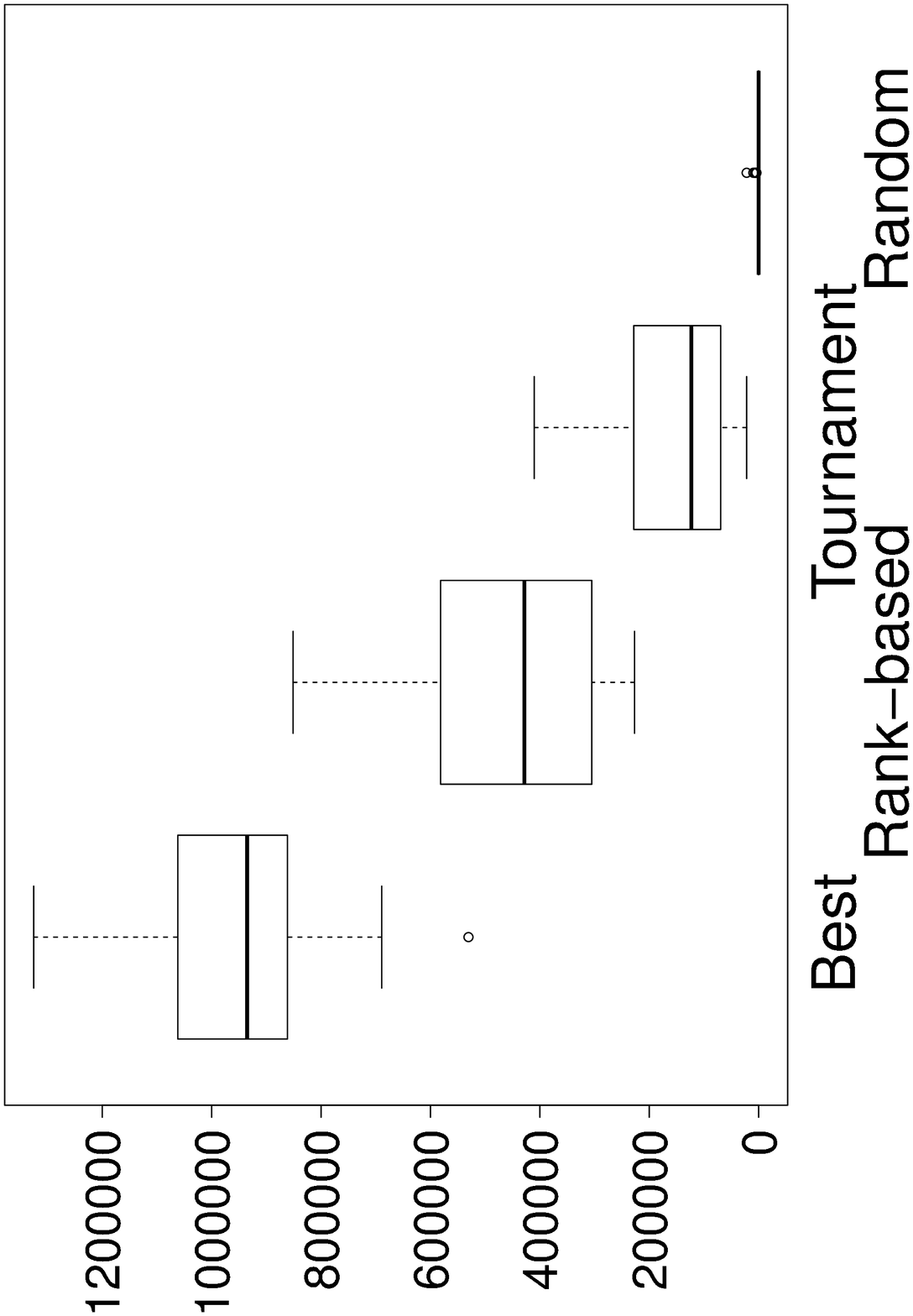}}
  \vspace{-3mm}
  \caption{\label{fig:navigMetrics}Box and whisker plots ($30$
    independent runs) of the comparison measures for the four selection
    methods on the navigation task. From top to bottom and left to
    right: $f_c$, $f_b$, $g_f$ and $f_a$. The label $p>0.01$ indicates no
    statistical difference for the corresponding two selection methods.}
\end{figure}

\begin{figure}[!h]
  \centering
  \resizebox{.49\linewidth}{!}{
    \includegraphics[width=\textwidth,angle=-90,origin=c]{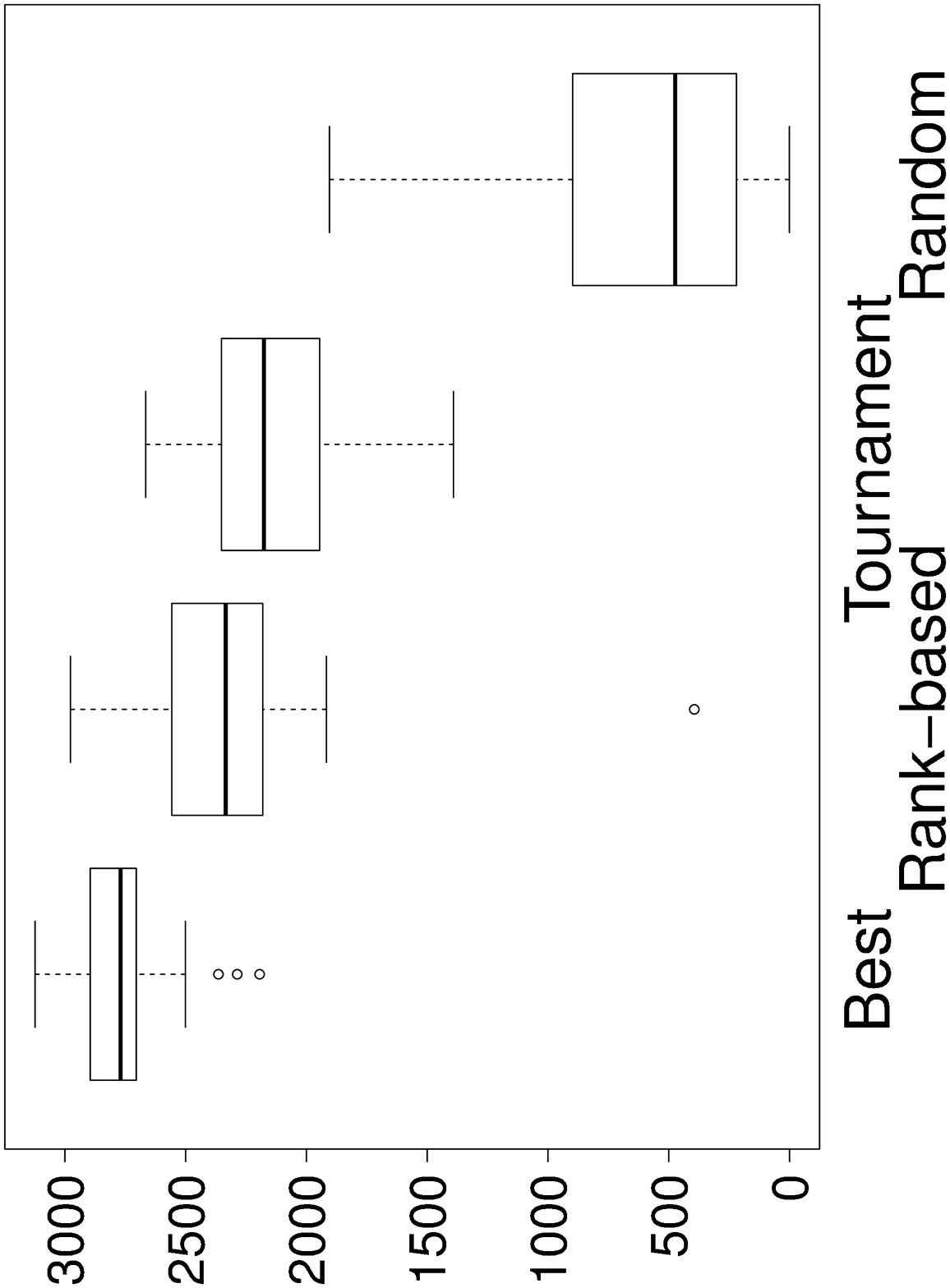}}
  \resizebox{.49\linewidth}{!}{
    \includegraphics[width=\textwidth,angle=-90,origin=c]{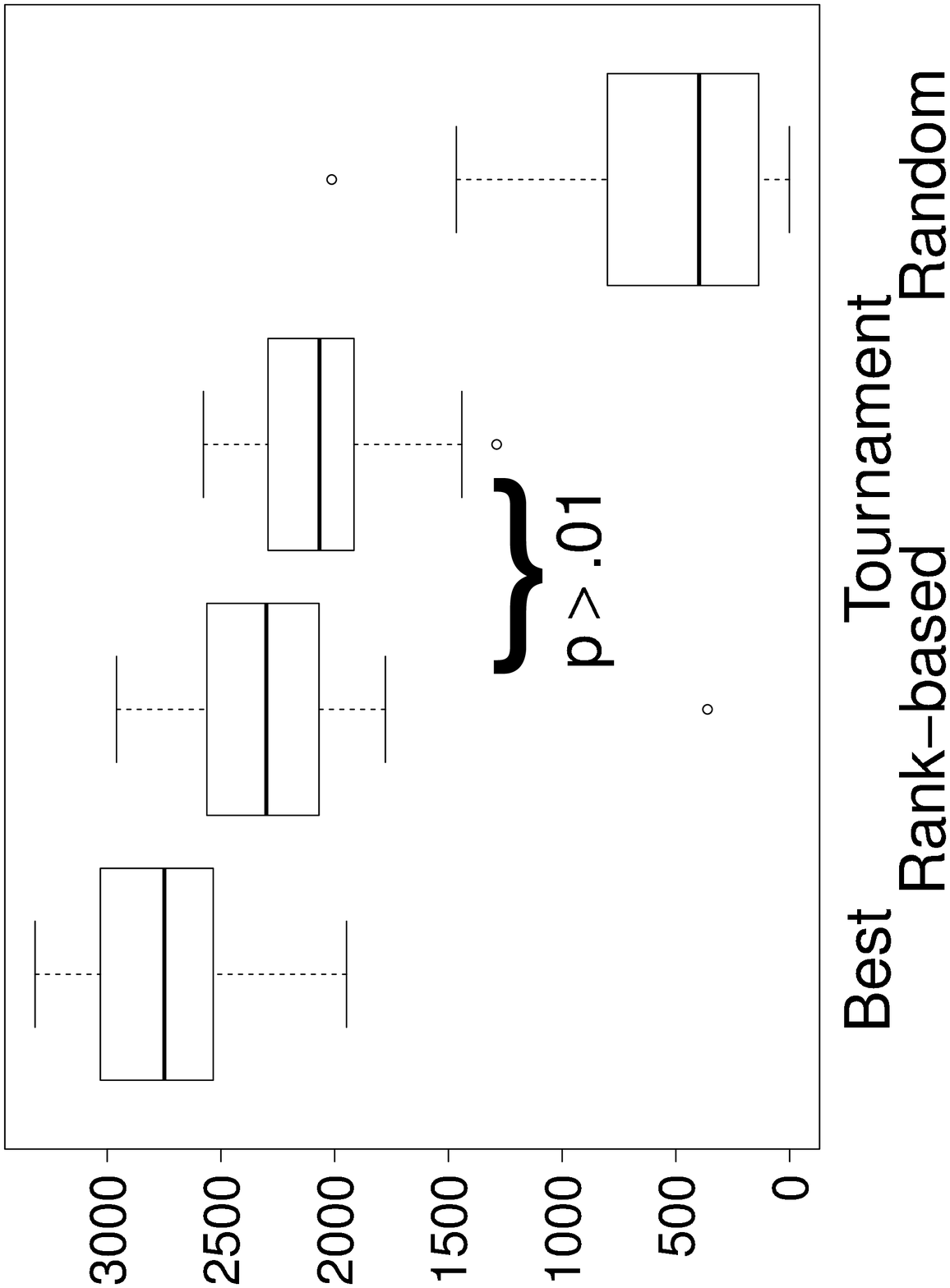}}
  \resizebox{.49\linewidth}{!}{
    \includegraphics[width=\textwidth,angle=-90,origin=c]{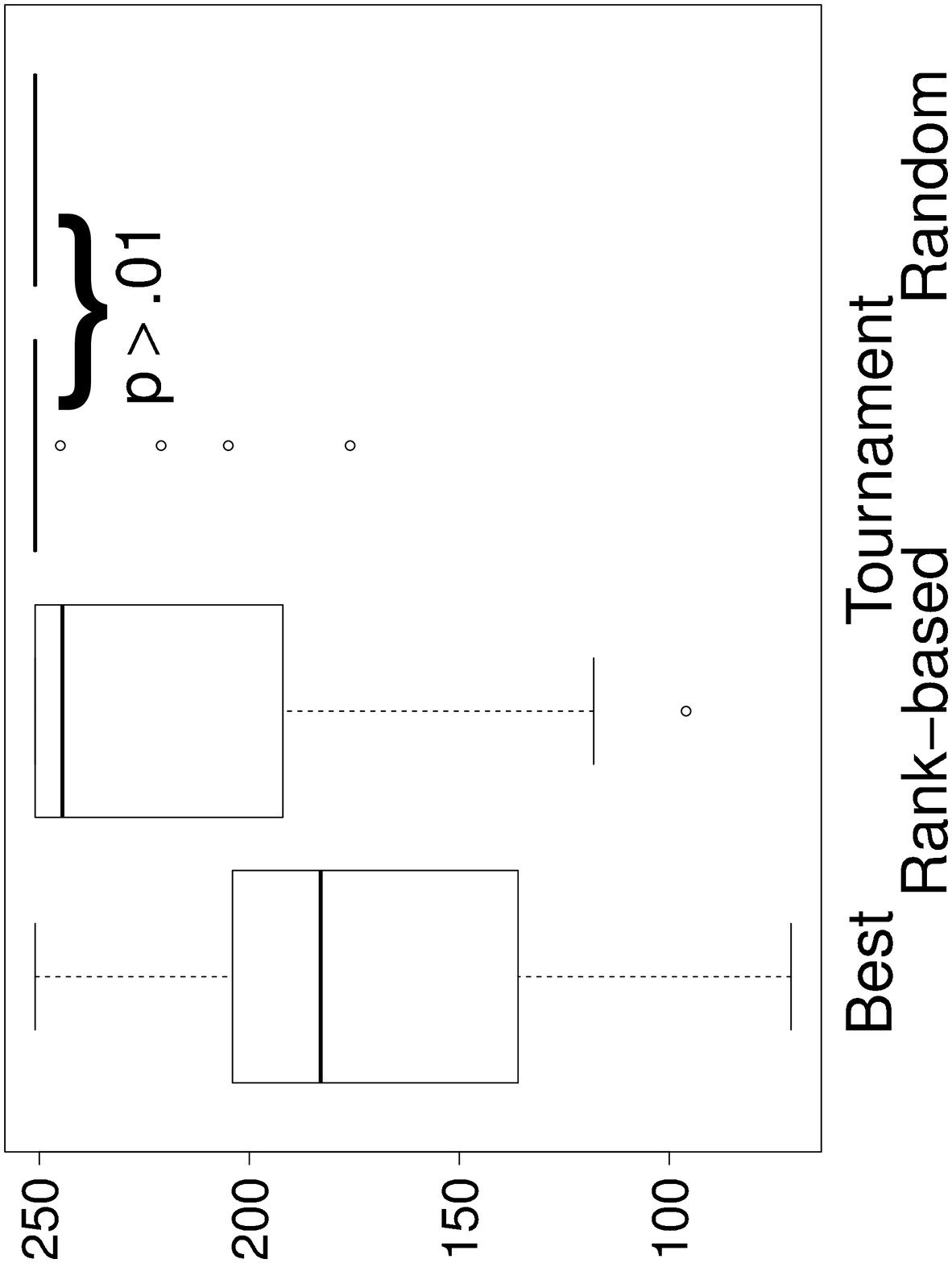}}
  \resizebox{.49\linewidth}{!}{
    \includegraphics[width=\textwidth,angle=-90,origin=c]{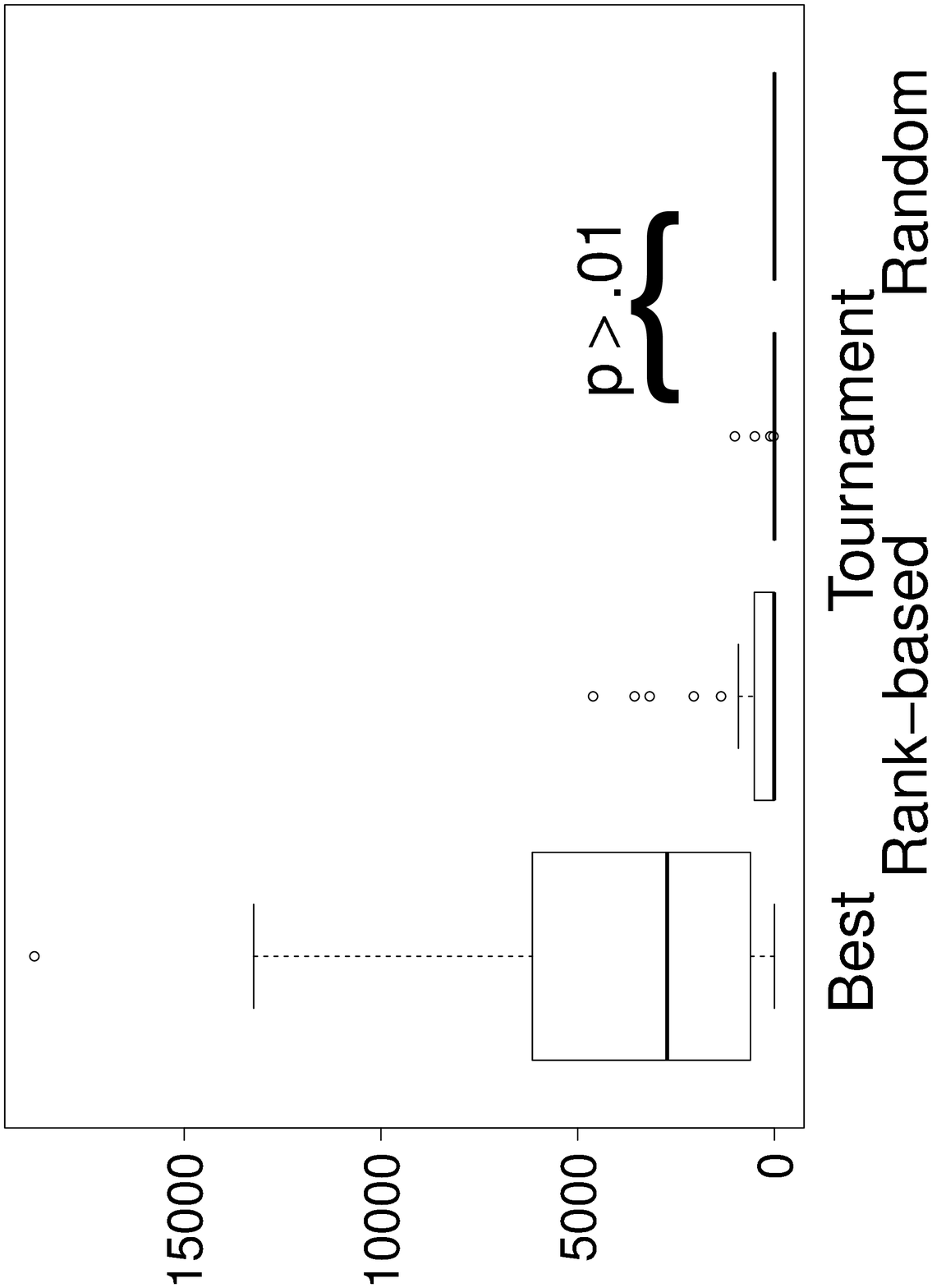}}
  \vspace{-3mm}
  \caption{\label{fig:foragMetrics} Box and whisker plots ($30$
    independent runs) of the comparison measures for the four selection
    methods on the foraging task. From top to bottom and left to
    right: $f_c$, $f_b$, $g_f$ and $f_a$. The label $p>0.01$ indicates no
    statistical difference for the corresponding two selection methods.}
\end{figure}

For both navigation and foraging tasks, we run $30$ independent runs
for each selection method, and we measured $F_s$ at each generation in
all runs. Figures~\ref{fig:navigEvo} and~\ref{fig:foragEvo} show the
median $F_s$ per generation over the $30$ runs for each task. We
computed the four performance metrics in the case of navigation
(Figure~\ref{fig:navigMetrics}) and of foraging
(Figure~\ref{fig:foragMetrics}). For both tasks, we performed
pairwise\footnote{Pairwise in this context means all combinations of
  pairs of selection methods, six combinations in our case.}
Mann-Whitney tests at $99\%$ confidence on these measures, between the
four selection methods.

On the one hand, upon analysis of Figure~\ref{fig:navigEvo} and
Figure~\ref{fig:foragEvo}, we observe that the swarm rapidly reaches a
high fitness level in both tasks whenever there is a task-driven
selection pressure, \ie with \textit{Best}, \textit{Rank-based} or
\textit{Binary tournament} selection. On the other hand, without any
selection pressure (\textit{Random}), learning is much
slower. Furthermore, for the three former selection methods the
algorithm reaches comparable levels of performance in terms of median
values of the swarm fitness. An exception can be noted for
\textit{Best} selection in the foraging task, which outperforms
\textit{Rank-based} and \textit{Binary tournament}.

Despite the lower performances achieved by \textit{Random}, the swarm
still manages to learn behaviors for both tasks. This can be seen in
the increasing trend of the median swarm fitness in
Figure~\ref{fig:navigEvo} and Figure~\ref{fig:foragEvo}.  This result
is expected on the navigation task. As it is the case in 
(\cite{medea2010}), environmental pressure drives evolution toward
behaviors that maximize mating opportunities and thus behaviors that
explore the environment, increasing the swarm fitness.

The same trend is also observed on the foraging task.  The improvement
is slower but still present with \textit{Random} selection. This
could be explained by the fact that collecting items is a byproduct of
maximizing mating opportunities. Agents collect items by chance while
they navigate trying to mate. When inspecting the swarm in the
simulator, we observed that, when selection pressure is present, the
evolved behaviors drive the agents toward food items which means
that the food sensors are in fact exploited. In other words, evolution
drove the controllers to use these sensors. However, without any
selection pressure (\textit{Random}), there can not be a similar
drive. We also observed this in the simulator: agents were not
attracted by food items for \textit{Random} selection.

When we analyze the comparison measures we introduced earlier, similar
trends are observed. Figure~\ref{fig:navigMetrics} (respectively
Figure~\ref{fig:foragMetrics}) shows the box and whiskers plots of the
four measures for each selection method over the $30$ runs for the
navigation task (respectively the foraging task).

On the navigation task, the pairwise comparisons of the four measures,
using Mann-Whitney tests at $99\%$ confidence level, yield significant
statistical difference between all selection methods except between
\textit{Best} and \textit{Rank-based} (p-value=$0.0795$) and between
\textit{Rank-based} and \textit{Binary tournament} (p-value $0.0116$)
in the case of the time to reach target ($g_f$).

We also observe that \textit{Best} reaches a higher swarm fitness for
the fixed budget than the rest of selection methods, and this level is
maintained at the end of evolution, as is shown in $f_c$ and $f_b$
(upper left and right in the figure). The target fitness level is
rapidly reached for the three methods inducing selection pressure, and
there is not significant difference between \textit{Best} and
\textit{Rank-based}, nor between \textit{Rank-based} and
\textit{Binary Tournament} w.r.t $g_f$ (lower left). Furthermore, in
the case of \textit{Best}, the required level is not only reached but
surpassed during the entire evolution, leading to a value of $f_a$
much higher than the ones of the rest of selection methods (lower
right). However, this is not the case for \textit{Random} selection
that has much lower $f_b$ (upper right) and $f_c$ (upper left), and
does not reach the target fitness level on more than half the runs
that were launched (lower left and right).

On the foraging task, there is a significant difference for all
pairwise comparisons, except between \textit{Binary Tournament} and
\textit{Random} in the case of the time to reach target, $g_f$, and
the accumulated fitness above target, $f_a$, (p-value=$0.0419$ in both
cases). This is explained by the fact that very few runs attained the
target fitness\footnote{For both tasks, the target fitness is $80\%$
  of the highest fitness reached by all methods during all runs.} in
which case $g_f$ is the last generation and $f_a$ is almost
zero. There is also no statistical difference between
\textit{Rank-based} and \textit{Binary Tournament} on the fixed budget
swarm fitness, $f_b$, (p-value=$0.0105$). This means that
\textit{Binary Tournament} reaches a fitness at the given budget that
is comparable to the one of \textit{Rank-based}, but it does not
maintain this level, since for these two methods the difference is
significant on $f_c$.

\textit{Best} also gives better results on the foraging task: a high
swarm fitness is reached and maintained at the end of evolution ($f_b$
and $f_c$, upper left and right). It surpasses the target fitness
level in almost all runs much faster and to a larger extent than
\textit{Rank-based}, that also manages to reach the required level for
most runs ($g_f$, lower left), although by a lower level ($f_a$, lower
right). This is not the case for \textit{Tournament} and
\textit{Random} that do not achieve the target fitness level for most
runs (lower left and right).

We can observe that all task-driven selection pressures yield much
better performances on both tasks compared to \textit{Random}
selection. Consequently, we may conclude that selection pressure has a
positive impact on performances, when solving a given task, and when
the objective is not only to achieve adaptation of the swarm as it was
the original motivation of mEDEA. Further, statistical tests show a
direct correlation between the selection pressure and the performances
achieved by the swarm on the two considered tasks. In other words, the
stronger the selection pressure, the better the performances reached
by the swarm.

In general, it has been argued that elitist strategies are not
desirable in traditional EA's, and the same argument holds for
traditional ER. This is due to the fact that elitist strategies may
lead to a premature convergence at local optima. There exists an
extensive body of work, especially in non-convex optimization, where
it is preferable to explicitly maintain a certain level of diversity
in the population to escape local optima and to deal with the
exploration vs. exploitation dilemma. This requirement is perhaps not
as strong in the context of distributed ER as our experiments
show. Selection is performed among a portion of the population at the
agent level, therefore, one might argue that these algorithms already
maintain a certain level of diversity inherent to the fact that
sub-populations are distributed on the different agents. Comparisons
with other approaches in which separated sub-populations are evolved,
such as spatially structured EA's (\cite{tomassini2005}) and island
models (\cite{alba2002}), could give further insights on the dynamics
of this kind of evolution.

%% file: fig/controller.pstex_t
\begin{picture}(0,0)%
\includegraphics{fig/controller.pstex}%
\end{picture}%
\setlength{\unitlength}{4144sp}%
\begingroup\makeatletter\ifx\SetFigFont\undefined%
\gdef\SetFigFont#1#2#3#4#5{%
  \reset@font\fontsize{#1}{#2pt}%
  \fontfamily{#3}\fontseries{#4}\fontshape{#5}%
  \selectfont}%
\fi\endgroup%
\begin{picture}(3088,4066)(1200,-4794)
\put(1351,-4741){\rotatebox{90.0}{\makebox(0,0)[lb]{\smash{{\SetFigFont{12}{14.4}{\rmdefault}{\mddefault}{\updefault}{\color[rgb]{0,0,0}bias}%
}}}}}
\put(1351,-3886){\rotatebox{90.0}{\makebox(0,0)[lb]{\smash{{\SetFigFont{12}{14.4}{\rmdefault}{\mddefault}{\updefault}{\color[rgb]{0,0,0}obstacle sensors}%
}}}}}
\put(1351,-1906){\rotatebox{90.0}{\makebox(0,0)[lb]{\smash{{\SetFigFont{12}{14.4}{\rmdefault}{\mddefault}{\updefault}{\color[rgb]{0,0,0}food sensors}%
}}}}}
\put(3961,-1996){\makebox(0,0)[lb]{\smash{{\SetFigFont{12}{14.4}{\rmdefault}{\mddefault}{\updefault}{\color[rgb]{0,0,0}$v_r$}%
}}}}
\put(3961,-3661){\makebox(0,0)[lb]{\smash{{\SetFigFont{12}{14.4}{\rmdefault}{\mddefault}{\updefault}{\color[rgb]{0,0,0}$v_t$}%
}}}}
\end{picture}%

%% file: fig/measure-1.pstex_t
\begin{picture}(0,0)%
\includegraphics{fig/measure-1.pstex}%
\end{picture}%
\setlength{\unitlength}{3947sp}%
\begingroup\makeatletter\ifx\SetFigFont\undefined%
\gdef\SetFigFont#1#2#3#4#5{%
  \reset@font\fontsize{#1}{#2pt}%
  \fontfamily{#3}\fontseries{#4}\fontshape{#5}%
  \selectfont}%
\fi\endgroup%
\begin{picture}(3234,2004)(679,-2203)
\put(826,-1261){\rotatebox{90.0}{\makebox(0,0)[lb]{\smash{{\SetFigFont{12}{14.4}{\familydefault}{\mddefault}{\updefault}{\color[rgb]{0,0,0}$f$}%
}}}}}
\put(3826,-2086){\makebox(0,0)[lb]{\smash{{\SetFigFont{12}{14.4}{\familydefault}{\mddefault}{\updefault}{\color[rgb]{0,0,0}$t$}%
}}}}
\put(2701,-2086){\makebox(0,0)[lb]{\smash{{\SetFigFont{12}{14.4}{\familydefault}{\mddefault}{\updefault}{\color[rgb]{0,0,0}$g_b$}%
}}}}
\put(3301,-1336){\makebox(0,0)[lb]{\smash{{\SetFigFont{12}{14.4}{\familydefault}{\mddefault}{\updefault}{\color[rgb]{0,0,0}$f_c$}%
}}}}
\end{picture}%

%% file: fig/measure-2.pstex_t
\begin{picture}(0,0)%
\includegraphics{fig/measure-2.pstex}%
\end{picture}%
\setlength{\unitlength}{3947sp}%
\begingroup\makeatletter\ifx\SetFigFont\undefined%
\gdef\SetFigFont#1#2#3#4#5{%
  \reset@font\fontsize{#1}{#2pt}%
  \fontfamily{#3}\fontseries{#4}\fontshape{#5}%
  \selectfont}%
\fi\endgroup%
\begin{picture}(3234,2004)(679,-2203)
\put(826,-1261){\rotatebox{90.0}{\makebox(0,0)[lb]{\smash{{\SetFigFont{12}{14.4}{\familydefault}{\mddefault}{\updefault}{\color[rgb]{0,0,0}$f$}%
}}}}}
\put(3826,-2086){\makebox(0,0)[lb]{\smash{{\SetFigFont{12}{14.4}{\familydefault}{\mddefault}{\updefault}{\color[rgb]{0,0,0}$t$}%
}}}}
\put(976,-511){\makebox(0,0)[lb]{\smash{{\SetFigFont{12}{14.4}{\familydefault}{\mddefault}{\updefault}{\color[rgb]{0,0,0}$f_b$}%
}}}}
\put(3076,-2086){\makebox(0,0)[lb]{\smash{{\SetFigFont{12}{14.4}{\familydefault}{\mddefault}{\updefault}{\color[rgb]{0,0,0}$g_b$}%
}}}}
\end{picture}%

%% file: fig/measure-3.pstex_t
\begin{picture}(0,0)%
\includegraphics{fig/measure-3.pstex}%
\end{picture}%
\setlength{\unitlength}{3947sp}%
\begingroup\makeatletter\ifx\SetFigFont\undefined%
\gdef\SetFigFont#1#2#3#4#5{%
  \reset@font\fontsize{#1}{#2pt}%
  \fontfamily{#3}\fontseries{#4}\fontshape{#5}%
  \selectfont}%
\fi\endgroup%
\begin{picture}(3234,2004)(679,-2203)
\put(826,-1261){\rotatebox{90.0}{\makebox(0,0)[lb]{\smash{{\SetFigFont{12}{14.4}{\familydefault}{\mddefault}{\updefault}{\color[rgb]{0,0,0}$f$}%
}}}}}
\put(3826,-2086){\makebox(0,0)[lb]{\smash{{\SetFigFont{12}{14.4}{\familydefault}{\mddefault}{\updefault}{\color[rgb]{0,0,0}$t$}%
}}}}
\put(2776,-2086){\makebox(0,0)[lb]{\smash{{\SetFigFont{12}{14.4}{\familydefault}{\mddefault}{\updefault}{\color[rgb]{0,0,0}$g_f$}%
}}}}
\put(976,-586){\makebox(0,0)[lb]{\smash{{\SetFigFont{12}{14.4}{\familydefault}{\mddefault}{\updefault}{\color[rgb]{0,0,0}$f_t$}%
}}}}
\end{picture}%

%% file: fig/measure-4.pstex_t
\begin{picture}(0,0)%
\includegraphics{fig/measure-4.pstex}%
\end{picture}%
\setlength{\unitlength}{3947sp}%
\begingroup\makeatletter\ifx\SetFigFont\undefined%
\gdef\SetFigFont#1#2#3#4#5{%
  \reset@font\fontsize{#1}{#2pt}%
  \fontfamily{#3}\fontseries{#4}\fontshape{#5}%
  \selectfont}%
\fi\endgroup%
\begin{picture}(3234,2004)(679,-2203)
\put(3826,-2086){\makebox(0,0)[lb]{\smash{{\SetFigFont{12}{14.4}{\familydefault}{\mddefault}{\updefault}{\color[rgb]{0,0,0}$t$}%
}}}}
\put(826,-1261){\rotatebox{90.0}{\makebox(0,0)[lb]{\smash{{\SetFigFont{12}{14.4}{\familydefault}{\mddefault}{\updefault}{\color[rgb]{0,0,0}$f$}%
}}}}}
\put(976,-586){\makebox(0,0)[lb]{\smash{{\SetFigFont{12}{14.4}{\familydefault}{\mddefault}{\updefault}{\color[rgb]{0,0,0}$f_t$}%
}}}}
\put(3001,-886){\makebox(0,0)[lb]{\smash{{\SetFigFont{12}{14.4}{\familydefault}{\mddefault}{\updefault}{\color[rgb]{0,0,0}$f_a$}%
}}}}
\end{picture}%

%% file: 05-conclusion.tex
\section{Conclusions}
In this paper, we studied the impact of task-driven selection
pressures in on-line distributed ER for swarm behavior learning. This
kind of algorithms raises several questions concerning the usefulness
of selection pressure (partial views of population, noisy fitness
values, etc.). We compared four selection methods inducing different
intensities of selection pressure on two tasks: navigation with
obstacle avoidance and collective foraging. Our experiments show that
selection pressure largely improves performances, and that the
intensity of the selection operator positively correlates with the
performances of the swarm.

Foraging and navigation can be considered as relatively simple tasks and
we believe that more complex and challenging ones, involving
deceptive fitness functions, could give further insights on selection
and evolution dynamics in the distributed case.

% Local IspellDict:        english